\documentclass[11pt,a4paper]{article}
\usepackage[hyperref]{acl2018}
\usepackage[utf8]{inputenc}
\usepackage{times}
\usepackage{latexsym}
\usepackage{graphicx}
\usepackage{amsmath}
\usepackage{multirow}
\usepackage{booktabs}
\usepackage{subfigure}
\usepackage[shortcuts]{extdash}
\usepackage{url}
\usepackage{comment}

\aclfinalcopy 

\title{Grammatical gender in Swedish is predictable \\ using recurrent neural networks}

\author{Edvin Listo Zec \\
  RISE Research Institutes of Sweden \\
  {\tt edvin.l.zec@gmail.com} \\\And
  Olof Mogren \\
  RISE Research Institutes of Sweden \\
  {\tt olof@mogren.one} \\}

\date{}

\begin{document}
\maketitle
\begin{abstract}
  The grammatical gender of Swedish nouns is a mystery.
  While there are few rules that can indicate the gender with some certainty, it does in general not depend on either meaning or the structure of the word.
  
  In this paper we demonstrate the surprising fact that grammatical gender for Swedish nouns can be predicted with high accuracy using a recurrent neural network (RNN) working on the raw character sequence of the word, without using any contextual information.
\end{abstract}

\section{Introduction}

There are two different grammatical genders for Swedish nouns, \textit{common (utrum)} and \textit{neuter (neutrum)} \cite{gullberg1954}. Swedish used to have more than two genders; however the masculine and the feminine later merged into the common gender and modern Swedish makes no difference between them. The grammatical gender affects the indefinite article ({\em en} boll -- {\em ett} bord; {\em English: a ball -- a table}), as well as the definite article and the definite suffix. The Institute for Language and Folklore, a Swedish  government agency, states that there is no unequivocal way of determining the grammatical gender of a word \cite{folklore}. Further the Swedish Academy states that the grammatical gender in Swedish is a lexical property which is to a large extent not dependent on neither meaning nor the structure of the noun \cite{teleman1999svenska}. There are exceptions: living things are generally utrum (the more common class), and there are a few specific suffixes that are reasonably good predictors.
However, determining the gender of a word is generally  considered a major difficulty for non-native Swedish speakers learning the language, and it is an open question how predictable it is.

In this paper, we investigate the predictability of grammatical gender for Swedish nouns without any context words.
We train a recurrent neural network (RNN) model to predict the property given only the raw character sequence as input.
The model quite easily learns the necessary patterns and achieves a 95\% accuracy on the test set, which should be compared to the $71\%$ majority class baseline.

\section{Prediction of grammatical gender using recurrent neural networks}


A recurrent neural network (RNN) is a feed forward artificial neural
network that can model a sequence of arbitrary length, using weight
sharing between each position in the sequence.
In the basic RNN variant, the transition function at time $t$ is a linear
transformation of the hidden state $\mathbf{h_{t-1}}$ and the input, followed by a
point-wise non-linearity:
\[ \mathbf{h}_t = \mbox{tanh}( W \mathbf{x}_t + U \mathbf{h}_{t-1} + \mathbf{b}),\]
where $W$ and $U$ are weight matrices, $\mathbf{b}$ is a bias vector, and tanh is
the selected nonlinearity.  $W$, $U$, and $\mathbf{b}$ are typically trained using some
variant of stochastic gradient descent (SGD).

Basic RNNs struggle with learning long-range dependencies and suffer from
the vanishing gradient problem.  This makes them difficult to
train~\cite{hochreiter1998vanishing,bengio1994learning}, and has
motivated the development of the Long short term memory
(LSTM) architecture~\cite{hochreiter1997lstm}, that to some extent solves these
shortcomings.  An LSTM is gated variant of the RNN where the cell at each step $t$
contains an internal memory vector $\mathbf{c}_t$, and three gates controlling
what parts of the internal memory will be kept (the forget
gate~$\mathbf{f}_t$), what parts of the input that will be stored in the
internal memory (the input gate~$\mathbf{i}_t$), as well as what will be
included in the output (the output gate~$\mathbf{o}_t$). Later, the gated recurrent network architecture (GRU) \cite{cho2014learning} was proposed, which only has two different gates.

\begin{figure*}[]
\centering
\includegraphics[width = 0.6\paperwidth]{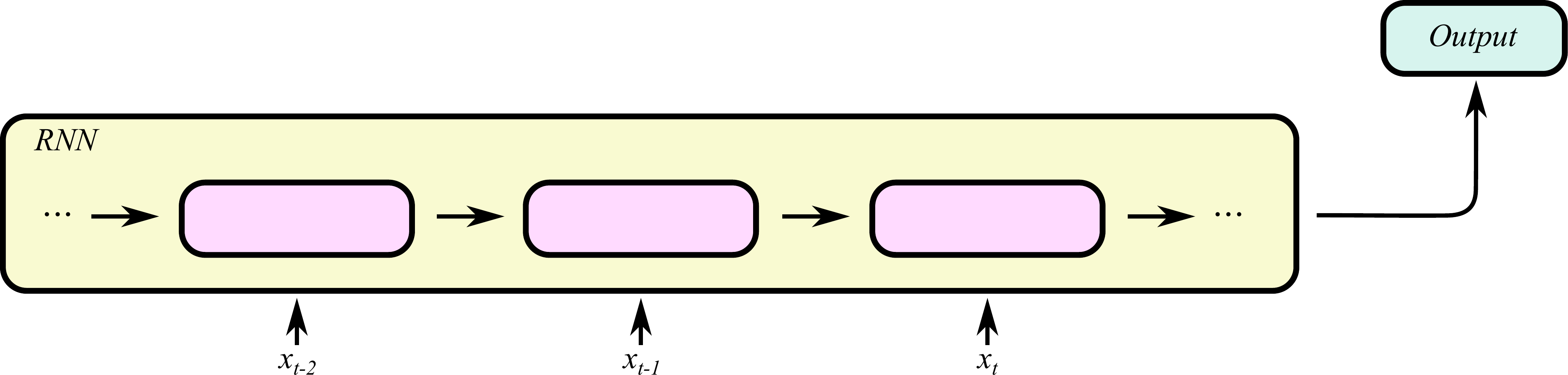}
\caption{The proposed character based RNN model used in this work. The output layer is a fully connected layer with one output unit and sigmoid actication function. }\label{fig:rnn-model}
\end{figure*}

In this paper, we propose a neural network model based on recurrent neural networks (see figure~\ref{fig:rnn-model}), that is trained to predict the grammatical gender of Swedish nouns.
The model has a fully connected output layer with one unit and a sigmoid activation function.

\section{Experimental setup}

Three main network architectures were set up and trained on the same dataset and then compared on the same test set. All words in the dataset were vectorised by transforming each character of the words into a unique integer. Then each vectorised word was padded with zeros at the end matching the length of the longest word, in our case 19.

We train, evaluate and compare three different models: a feed-forward network, an LSTM network and a GRU network. The first model consists of an embedding layer which takes as input each character sequence describing a word, and outputs an embedding of dimension 60. It is then followed by two dense layers with 128 and 1 unit(s) respectively for each layer, with a $\tanh$ activation in the hidden layer and a sigmoid in the output layer. The second model is composed of the same embedding layer, followed by an RNN with a GRU with 64 units and a $\tanh$ activation. The RNN outputs a whole sequence into a dense layer with one unit and the sigmoid as activation. The third model is the same as the second, with the difference being that it uses LSTM units instead of GRUs.





\subsection{Hyperparameters}
All models were trained with the Adam optimiser with the same hyperparameter settings as that of the paper \cite{kingma2014adam}, where they used a learning rate $\alpha = 0.001$, decay rates $\beta_1 = 0.9$ and $\beta_2 = 0.999$, and $\epsilon=10^{-8}$.


\subsection{Dataset}
The dataset consists of 88480 Swedish nouns acquired from SALDO \cite{borin2013saldo}\footnote{Språkbanken (the Swedish Language Bank): \url{https://spraakbanken.gu.se/swe/resurs/saldo}}, labelled with grammatical gender. According to the Swedish Academy about two thirds of Swedish nouns are utrum and one third neutrum \cite{teleman1999svenska}. In our dataset, $71\%$ of the words belong to the utrum class, and $29\%$ to the neutrum. Further, the Swedish Academy notes that there are some common suffixes that strongly correlate with the grammatical gender. For example, words ending with \textit{ing, tion, het} and \textit{ist} are mostly associated with utrum and words ending with \textit{eri, skop} and \textit{gram} are mostly associated with neutrum. In table \ref{tab:wordstat} suffixes correlated with a certain class a listed.

As it is well known that the definite suffix indicates the grammatical gender, and the signal can be picked up from plural suffixes too, only singular indefinite nouns are included in the dataset.

\begin{table}[h]
    \centering
    \begin{tabular}{c|c|c}
         Suffix & Number of occurrences & Fraction utrum  \\ \midrule
         het & 8180 & 0.9999 \\
         tion & 1194 & 0.9992 \\
         ist & 509 & 0.9980 \\
         ing & 9384 & 0.9960 \\
         are & 3062 & 0.9941 \\
         skop & 35 & 0.1429 \\
         eri & 416 & 0.0385 \\
         gram & 109 & 0.0183 \\
         ande & 8931 & 0.0076 \\
    \end{tabular}
    \caption{Occurrences of words in the dataset and the fraction of the utrum class.}
    \label{tab:wordstat}
\end{table}

\begin{table*}[]
\begin{tabular}{ll}
\textbf{Correctly predicted word}         & \textbf{Correct label} \\
\midrule
aortaruptur             & U             \\
utrikesfråga            & U             \\
elektroakustik          & U             \\
överhopande             & N             \\
aschle                  & N             \\
utvecklingsstöd         & N             \\
\end{tabular}
\begin{tabular}{ll}%
\textbf{Incorrectly predicted word}  & \textbf{Correct label} \\
\midrule
hut                   & U             \\
lyster                & U             \\
nukleon               & U             \\
drogsug               & N             \\
zeugma                & N             \\
mattesnille           & N             \\
\end{tabular}
\caption{Randomly selected words from the test set that the proposed model predicted correctly (left), and incorrectly (right), respectively. U=utrum, N=neutrum.}
\label{tab:example-predictions}
\end{table*}

\subsection{Training}
A batch size of 32 was used in our experiment for all models. Furthermore, all models were trained until no change was observed in the validation loss for 50 epochs.

\subsection{Evaluation}

The models were evaluated using prediction accuracy. The dataset was split up into $60\%$ training, $20\%$ validation and $20\%$ test. A second set was created from the test set, where we removed all words with the common \textit{ing, tion, het, ist} and \textit{eri} suffixes. We can thus see how well a model learns different words and how it performs without common suffixes to its disposal.
In the evaluation, the prediction accuracy is reported for the test set.

\section{Results}
In table \ref{tab:testacc} the results from the three different neural networks are presented, with the best performing network in bold. The best model was the LSTM, with a classification accuracy of $95.15\%$ on the test set and $92.68\%$ on the test set with removed words. In table \ref{tab:prec} precision, recall and F1-score for the LSTM model are shown.

In table \ref{tab:example-predictions}, we present a random selection of words for which the proposed model correctly predicts the gender, and a random selection of words where it fails.

\begin{figure*}[htp]
  \centering
  \subfigure[t-SNE visualisation of the test set.]{\includegraphics[scale=0.3]{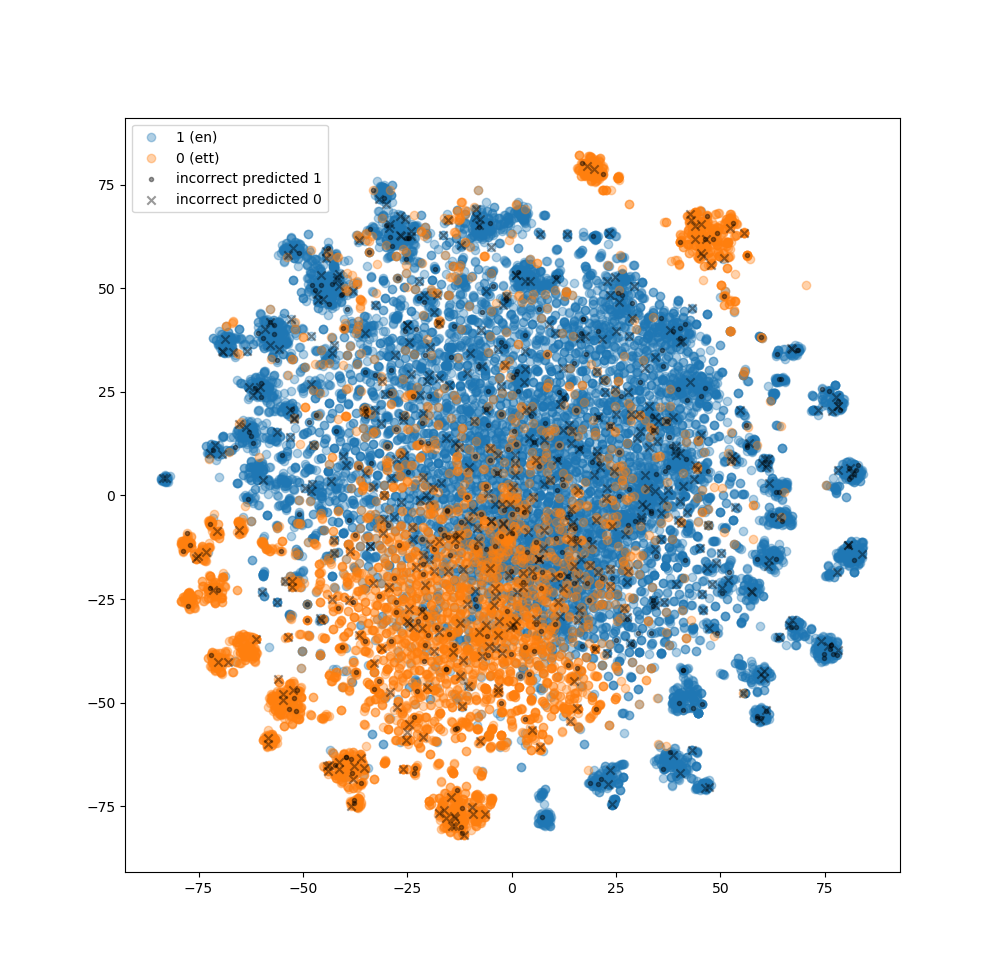}}
  \subfigure[t-SNE visualisation of the test set with removed suffixes.]{\includegraphics[scale=0.3]{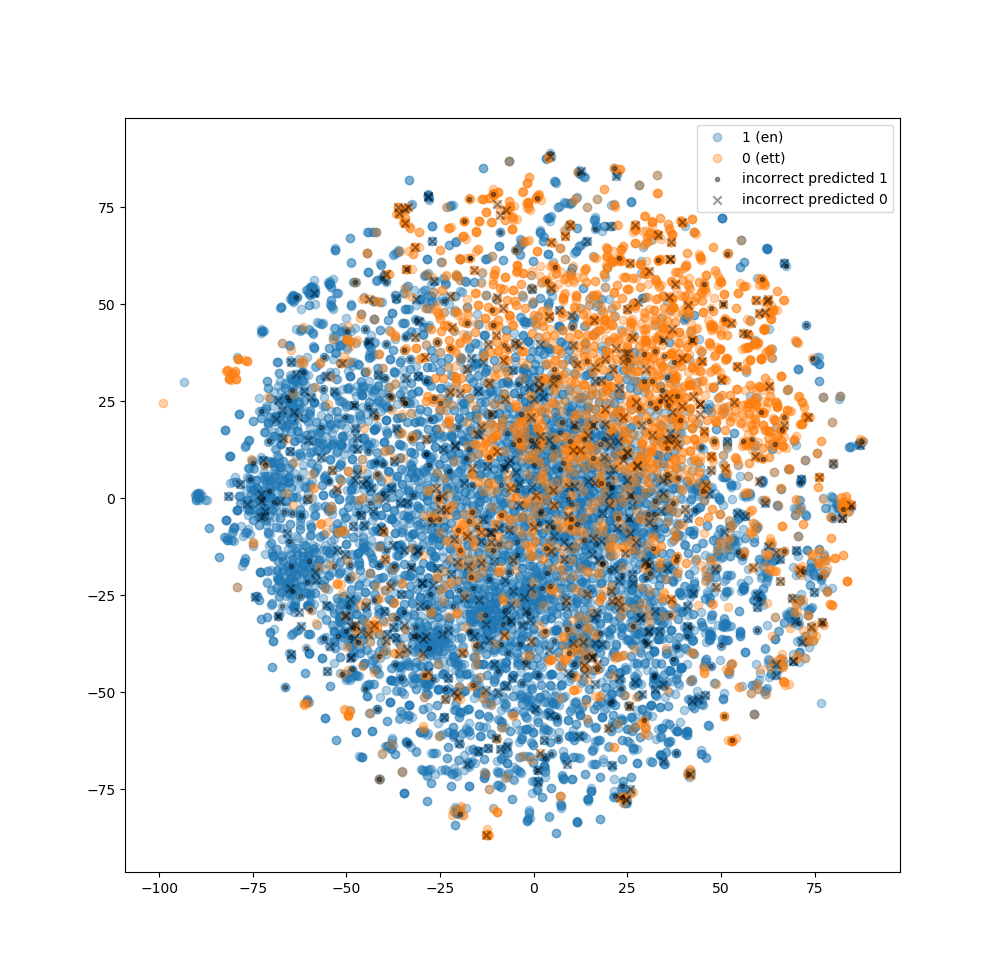}}
  \caption{t-SNE on the output from the LSTM layer with true labels and incorrect model predictions.}
  \label{tsne}
\end{figure*}

Figure \ref{tsne} depicts a t-SNE visualisation of the output from the LSTM layer on the two different test datasets.

\begin{table}[h!]
    \centering
    \begin{tabular}{c|c|c}
         & Test set & Test set with removed words  \\ \hline
        Dense & 0.8548 & 0.8577 \\ 
         GRU & 0.9492 &  0.9255 \\
          LSTM & 0.9515 & 0.9268 \\
          CNN & \textbf{0.9551} & \textbf{0.9324}
    \end{tabular}
    \caption{The accuracy for different test sets.}
    \label{tab:testacc}
\end{table}

\begin{table}[h!]
    \centering
    \begin{tabular}{c|c|c|c}
                 & Precision & Recall & F1-score \\ \hline
         Neutrum & 0.90 (0.85) & 0.93 (0.89) & 0.92 (0.87) \\
         Utrum & 0.97 (0.96) & 0.96 (0.94) & 0.97 (0.95)
    \end{tabular}
    \caption{Precision, recall and F1-score for the LSTM model on the test set. Test set with removed suffixes in parenthesis.}
    \label{tab:prec}
\end{table}

\section{Related work}

Character-based recurrent neural networks have demonstrated effective for identifying word forms and inflectional patterns.
In morphological relational reasoning, they can be used to generate a word in a form as demonstrated by another inflected word \cite{mogren2017character}.
They are also the state-of-the-art in morphological inflection and reinflection \cite{cotterell2016sigmorphon,kann2016med}.
In the current work, we instead focus on a single property that we want to infer from the raw character sequence of a given word. This means that we have a smaller model and a shorter path from the learning objective to backpropagate through.
The patterns that we do learn can however be similar to those in the works mentioned above.

\cite{nastase2009s} trained a \textit{kernel ridge regression (KRR)} model to predict the gender of nouns in German and Romanian to a high precision. It is however well known that Romanian has phonological patterns that indicate gender, while German is more similar to Swedish, in that there are a number of rules that have exceptions. Both German and Romanian have three genders each, while Swedish has only two.

\cite{desai2015staged} used a morphological analyzer to extract features to train a classifier for the grammatical gender in Konkani.
\cite{dinu2012romanian} trained machine (SVM) using ngram features and finds support for Romanian being a two-gender language, rather than the normally considered three genders.

\cite{cucerzan2003minimally} trained trie models that depend on morphological analysis (suffix cues) and contextual information from a corpus to determine the gender of a word. The model is bootstrapped with a small list of tagged words. The system was evaluated on French, Romanian, Slovene, Spanish, and Swedish. 
They conclude that a one-character suffix is too short to be a good predictor for Swedish. They end up achieving good prediction accuracy.
However, they learn the gender based on the context of nouns in a corpus, which is a different task than the one we are solving.

\cite{basirat2018linguistic} use word embeddings to predict the grammatical gender of Swedish nouns. Here, context of nouns is also used and they are able to achieve a high classification accuracy of $93.7\%$.

In the current work, we are not depending on any such corpus; the proposed model works solely using the raw character stream of the word. Thus, we use much less information as compared to context based methods.



\section{Discussion and conclusions}

In this work, we have seen that a simple RNN-based model is able to predict grammatical gender in Swedish nouns, a lexical property that is considered difficult and does not adhere to rules other than in a few special cases.
In contrast, our results show that we can learn the necessary patterns only by looking at the sequence of characters in a given word, without using any context words.





\bibliography{bibliography}
\bibliographystyle{acl_natbib}

\appendix
\section{Appendix} \label{appendix}

\begin{table}[]
\begin{tabular}{ll}
\textbf{Correctly predicted word}         & \textbf{Correct label} \\
aortaruptur             & 1             \\
utrikesfråga            & 1             \\
nappa                   & 1             \\
djurvänlighet           & 1             \\
silverklang             & 1             \\
ukrainska               & 1             \\
förlöpare               & 1             \\
befästande              & 0             \\
förhydande              & 0             \\
projekterbarhet         & 1             \\
elektroakustik          & 1             \\
överhopande             & 0             \\
aschle                  & 0             \\
utvecklingsstöd         & 0             \\
amoralitet              & 1             \\
fostrare                & 1             \\
catering                & 1             \\
omotsäglighet           & 1             \\
ovilja                  & 1             \\
rättssal                & 1             \\
utpekning               & 1             \\
otrohetsskandal         & 1             \\
kyrkostadga             & 1             \\
dagboksskrivande        & 0             \\
överlackerbarhet        & 1             \\
pressveck               & 0             \\
vattenled               & 1             \\
urklipp                 & 0             \\
rymd                    & 1             \\
pressgurka              & 1             \\
altarkärl               & 0             \\
glav                    & 1             \\
stjärngranne            & 1             \\
bordsskiva              & 1             \\
desinficeringsmedel     & 0             \\
svedenborgare           & 1             \\
krigsdans               & 1             \\
upphävande              & 0             \\
meningsmotståndare      & 1             \\
oapplåderbarhet         & 1             \\
opromulgerbarhet        & 1             \\
stoftpartikel           & 1             \\
ignorerbarhet           & 1             \\
sammetsblomster         & 0             \\
stångryttare            & 1             \\
rafräschissör           & 1             \\
byrå                    & 1             \\
naturlighet             & 1             \\
porrtidning             & 1             \\
hjässa                  & 1             \\
\end{tabular}
\caption{Randomly selected words that the proposed model predicted correct.}
\end{table}

\begin{table}[]
\begin{tabular}{ll}
\textbf{Incorrectly predicted word}  & \textbf{Correct label} \\
hut                   & 1             \\
bomullstuss           & 1             \\
lyster                & 1             \\
drogsug               & 0             \\
nukleon               & 1             \\
zeugma                & 0             \\
självdisciplin        & 1             \\
tjänstenit            & 0             \\
mörkalv               & 1             \\
avföringstransplantat & 0             \\
particip              & 0             \\
älgko                 & 1             \\
guckusko              & 1             \\
vedlår                & 1             \\
decimalkomma          & 0             \\
hälsodryck            & 1             \\
luftled               & 1             \\
synkop                & 1             \\
finne                 & 1             \\
näst                  & 0             \\
underhållskolonn      & 1             \\
jättetyfon            & 1             \\
tågstopp              & 0             \\
stråt                 & 1             \\
riboflavin            & 1             \\
maraton               & 0             \\
fortuna               & 0             \\
flusspat              & 1             \\
barr                  & 1             \\
mattesnille           & 0             \\
berättarjag           & 0             \\
tussilago             & 1             \\
eurodollar            & 1             \\
perm                  & 0             \\
kundnöjdhetsenkät     & 1             \\
chassi                & 0             \\
gen                   & 1             \\
björnbindsle          & 0             \\
kartotek              & 0             \\
cyklop                & 1             \\
akantom               & 0             \\
dego                  & 1             \\
cocktailparty         & 0             \\
sudd                  & 0             \\
dödsur                & 0             \\
schapp                & 0             \\
mc-gäng               & 0             \\
stins                 & 1             \\
bukfett               & 0             \\
arbetshandikapp       & 0             
\end{tabular}
\caption{Randomly selected words that the proposed model predicted wrong.}
\end{table}

\end{document}